\documentclass[sigconf]{acmart}
\renewcommand\footnotetextcopyrightpermission[1]{} 
\AtBeginDocument{%
  }
\usepackage{caption}
\copyrightyear{2018}
\acmYear{2018}
\acmDOI{XXXXXXX.XXXXXXX}
\acmISBN{978-1-4503-XXXX-X/2018/06}

\acmSubmissionID{8727}

\usepackage{multirow}
\usepackage{tabularx}
\begin{document}

\title{RACANet: Reliability-Aware Crowd Anchor Network for RGB-T Crowd Counting}
\author{Jinghao Shi}
\affiliation{%
  \institution{School of Computer Science, China University of Geosciences, Wuhan}
  \city{Wuhan}
  \country{China}
}

\author{Mengqi Lei}
\affiliation{%
  \institution{School of Software, Tsinghua University}
  \city{Beijing}
  \country{China}
}

\author{Kunliang He}
\affiliation{%
  \institution{School of Computer Science, China University of Geosciences, Wuhan}
  \city{Wuhan}
  \country{China}
}

\author{Yun Li}
\affiliation{%
  \institution{School of Computer Science, China University of Geosciences, Wuhan}
  \city{Wuhan}
  \country{China}
}

\author{Wei Bao}
\affiliation{%
  \institution{School of Software, Tsinghua University}
  \city{Beijing}
  \country{China}
}

\author{Siqi Li}
\affiliation{%
  \institution{School of Software, Tsinghua University}
  \city{Beijing}
  \country{China}
}

\renewcommand{\shortauthors}{Shi et al.}



\begin{abstract}
RGB-Thermal (T) crowd counting aims to integrate visible-spectrum and thermal infrared information to improve the robustness of crowd density estimation in complex scenes. Although existing studies generally improve counting accuracy through cross-modal feature fusion, most current methods rely on implicit cross-modal fusion strategies and lack explicit modeling of local spatial discrepancies as well as fine-grained characterization of modality reliability at the positional level, thereby limiting the accuracy and interpretability of the fusion process. To address these issues, this paper proposes a two-stage fusion framework, RACANet, a Reliability-Aware Crowd Anchor Network for RGB-T crowd counting. First, we introduce a lightweight cross-modal alignment pretraining stage, which explicitly learns cross-modal semantic correspondences through crowd-prior supervision and local bidirectional soft matching. Then, based on the priors learned during pretraining, a Local Anchor Fusion Module (LAFM) is introduced in the formal training stage. This module generates local semantic anchors by aggregating features from highly reliable regions and further enables adaptive pixel-level feature redistribution with a local attention mechanism. In addition, we propose a discrepancy-aware consistency constraint to dynamically coordinate the reliability of regions where modal representations are consistent. Experiments conducted on two widely used benchmark datasets, RGBT-CC and Drone-RGBT, demonstrate that RACANet outperforms existing methods. The anonymous code is available at \url{https://anonymous.4open.science/r/RACANet-9985}.
\vspace{-4pt}
\end{abstract}



\keywords{RGB-Thermal, Crowd Counting, Cross-Modal Pretraining, Modality Reliability Modeling, Local Anchor Fusion}

\addtolength{\abovedisplayskip}{-5pt}
\addtolength{\belowdisplayskip}{-3pt}
\addtolength{\abovedisplayshortskip}{-5pt}
\addtolength{\belowdisplayshortskip}{-3pt}



\maketitle
\pagestyle{plain}
\vspace{-6pt}
\section{Introduction}
\begin{figure*}[t]
    \centering
    \includegraphics[width=\textwidth]{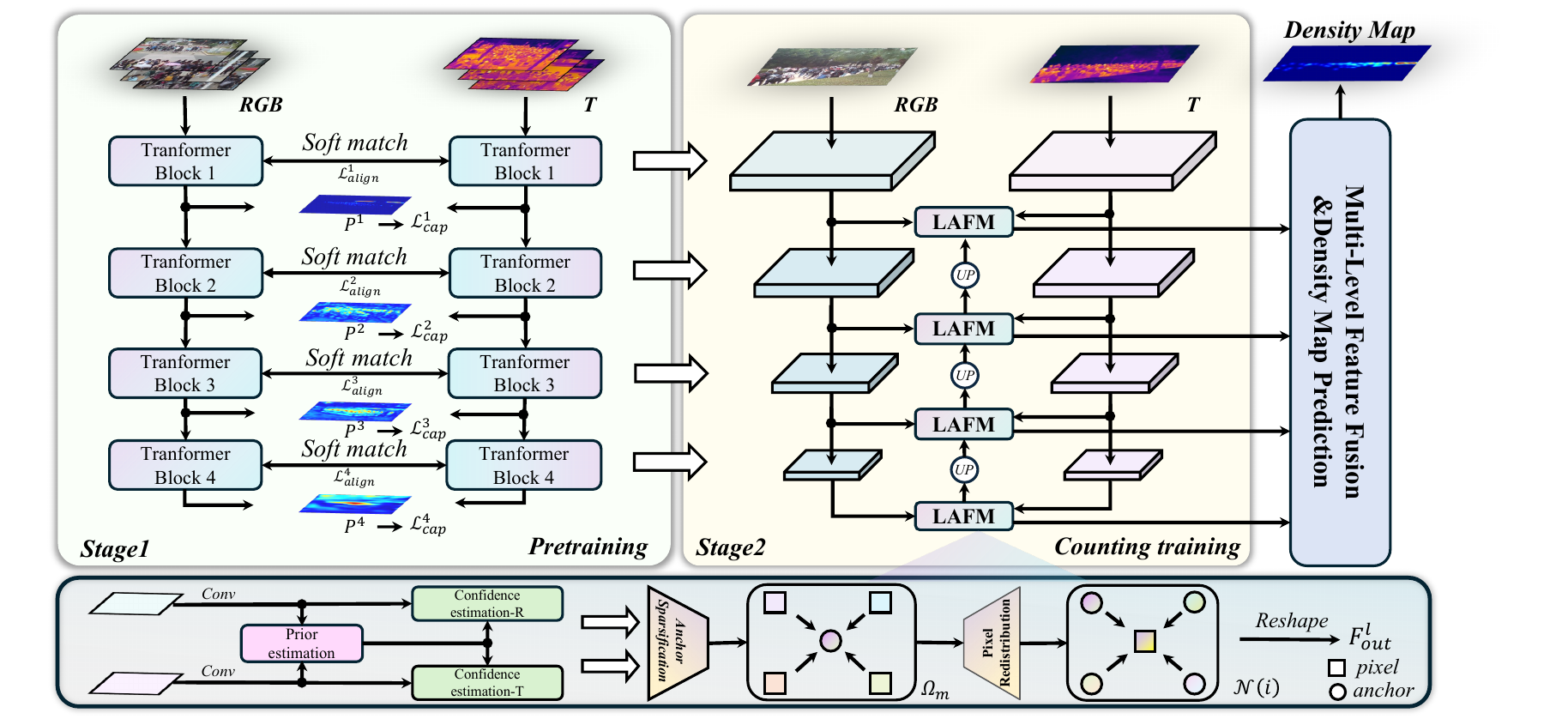}
    \vspace{-6mm}
    \captionsetup{font=footnotesize}
    \caption{The RACANet framework. Stage 1: A dual-branch PVTv2 backbone is adopted to perform soft feature matching and generate crowd-aware priors. Stage 2: The LAFM is introduced into multi-scale features, and the resulting fused features are then fed into the decoder. The bottom panel illustrates the workflow of this module: modality reliability is first estimated based on the crowd-aware priors, after which anchor aggregation and pixel-level semantic redistribution are performed within local regions.}
    \label{fig:framework}
    
    \vspace{-3.1mm}
\end{figure*}

Crowd counting has important application value in public safety, urban planning, and video surveillance ~\cite{crowdcountingsurvey,bmcc}. However, traditional counting methods that rely solely on visible-light images often suffer from severe performance degradation when confronted with complex open scenarios, such as sudden illumination changes or adverse weather conditions ~\cite{mmccn,rgbtcc,dong2025interleaved}. To overcome the limitations of a single sensor, RGB-T crowd counting, which integrates visible-light and thermal infrared information, has become a key research direction for improving crowd density estimation in challenging scenes.

Although the introduction of the thermal infrared modality provides rich complementary information, achieving effective fusion of the two modalities remains highly challenging. Due to physical differences in camera parallax and imaging mechanisms, RGB and thermal images can usually only achieve coarse spatial alignment, and local positional offsets are therefore inevitable~\cite{csanet,shi2025bicor}. Moreover, modality quality is not uniform across the same spatial locations: under conditions such as strong illumination, occlusion, or thermal interference, the signal quality of RGB and thermal modalities can fluctuate significantly~\cite{cginet,bgdfnet,dong2024novel}. Existing methods, although they improve performance through feature concatenation or attention mechanisms, mostly rely on implicit fusion paradigms~\cite{yinshifusion} and often assume that the two modalities are sufficiently aligned within local regions, while lacking explicit modeling of local cross-sensor deviations as well as fine-grained characterization of position-level modality reliability~\cite{misfnet,yolov13}, which severely limits the accuracy and interpretability of cross-modal fusion.

To address these challenges, we propose RACANet, a reliability-aware crowd anchor network for RGB-T crowd counting. Different from direct end-to-end learning, RACANet introduces a lightweight cross-modal alignment pretraining stage. Through crowd-aware prior supervision and bidirectional soft matching within local windows, the network is guided to explicitly learn local cross-modal correspondences. Based on this structured prior, we further propose a local anchor fusion module (LAFM) guided by the crowd-aware priors learned in the pretraining stage. Within local neighborhoods, LAFM independently evaluates pixel-level reliability for the two modalities, and achieves highly interpretable and low-complexity feature fusion through sparse anchor extraction and adaptive semantic redistribution. Meanwhile, we introduce a discrepancy-aware consistency constraint, which encourages reliability coordination in regions where cross-modal representations are consistent, while preserving necessary modality specificity in regions with significant discrepancies. We conduct comprehensive experiments on two widely used RGB-T datasets to demonstrate the performance of RACANet and provide an in-depth analysis of the experimental results.
The main contributions of this paper are summarized as follows:
\vspace{-3pt}
\begin{itemize}
    \item We propose a two-stage RGB-T crowd counting framework, RACANet. By introducing a lightweight cross-modal alignment pretraining stage, the network is able to explicitly learn local spatial correspondences and provide stable crowd-prior guidance for subsequent fusion.

    \item We design a local anchor fusion module (LAFM) and a discrepancy-aware consistency constraint. Through local prototype aggregation and pixel-wise semantic redistribution, the proposed method achieves fine-grained modality reliability estimation and cross-modal feature fusion with low computational overhead.
    
    \item We conduct extensive experiments on two widely used public benchmarks, RGBT-CC and Drone-RGBT. The results demonstrate that RACANet significantly outperforms existing state-of-the-art methods in counting accuracy.
    \vspace{-3pt}
\end{itemize}

\vspace{-3pt}
\section{Related Work}

\subsection{RGB-T Crowd Counting}
As the thermal infrared modality has been introduced into crowd analysis under complex illumination and adverse environmental conditions, RGB-T crowd counting has gradually emerged as an independent research direction. Peng et al. constructed the Drone-RGBT dataset and proposed MMCCN~\cite{mmccn}, while Liu et al. further released the large-scale RGBT-CC~\cite{rgbtcc} benchmark and proposed a cross-modal collaborative representation framework, thereby validating the effectiveness of collaborative multimodal representation learning for density regression tasks. Building on these efforts, subsequent studies have mainly focused on enhancing cross-modal interaction and feature aggregation. CSA-Net~\cite{csanet} addresses scale variations through scale-aware cross-modal aggregation and a channel attention mechanism. CGINet~\cite{cginet} and MC3Net~\cite{mc3net} emphasize bidirectional interaction, guidance, compensation, and collaboration between modalities, while BGDFNet~\cite{bgdfnet} and MISF-Net~\cite{misfnet} further improve fusion performance through gated dynamic interaction and modality-specific decomposition mechanisms. Existing RGB-T crowd counting methods have demonstrated the effectiveness of bimodal complementarity. However, the prevailing paradigm typically relies on global interactions to strengthen fusion, while lacking explicit modeling of local cross-sensor discrepancies and rarely characterizing reliability differences between modalities at the positional level.

\subsection{Cross-Modal Alignment and Reliability-Aware Fusion}
Recent studies have gradually recognized that modality misalignment itself is a core bottleneck constraining fusion quality~\cite{dong2024novel,bgdfnet,lei2025softhgnn}. In the field of RGB-T crowd counting, CrowdAlign~\cite{crowdalign} addresses the modality misalignment problem by proposing a shared-weight dual-layer alignment fusion strategy. BMCC~\cite{bmcc} transforms bimodal learning into tri-modal collaborative modeling. C4-MIM~\cite{c4mim} explicitly exploits the shared information across modalities through mutual information maximization. CFAF-Net~\cite{cfafnet} alleviates the spatial offsets between RGB and thermal infrared modalities through a frequency-cascade fusion strategy. These advances indicate that RGB-T crowd counting research has begun to shift from merely enhancing cross-modal interaction toward simultaneously addressing misalignment robustness and modality reliability.

This trend is equally evident in broader multimodal vision tasks. In multispectral pedestrian detection, IAF R-CNN~\cite{iafrcnn} and GAFF~\cite{gaff} adaptively adjust modality fusion weights according to illumination conditions. In RGB-T salient object detection, DCNet~\cite{dcnet} and SACNet~\cite{sacnet} are among the first to explicitly model cross-modal correlations and perform asymmetric association learning for weakly aligned scenarios. In object tracking, methods such as QAT~\cite{qat} have started to directly learn reliability weights for different modalities through supervised learning.

Recent studies suggest that when strict pixel-level alignment is difficult to achieve, the key to cross-modal modeling does not lie in forcibly sharing spatial positions, but rather in learning asymmetric cross-modal correspondences within local regions and identifying the relative reliability of different modalities at different locations\cite{crowdalign,ddranet}. However, in RGB-T crowd counting, these two capabilities are still often implicitly embedded within a unified fusion module, lacking both explicit local correspondence learning tailored to crowd regions and interpretable position-level reliability characterization. To address this limitation, this paper further explicitly integrates local alignment priors with position-level reliability modeling, thereby enabling more robust and more interpretable cross-modal fusion.
\begin{figure*}[t]
    \centering
    \includegraphics[width=0.95\linewidth]{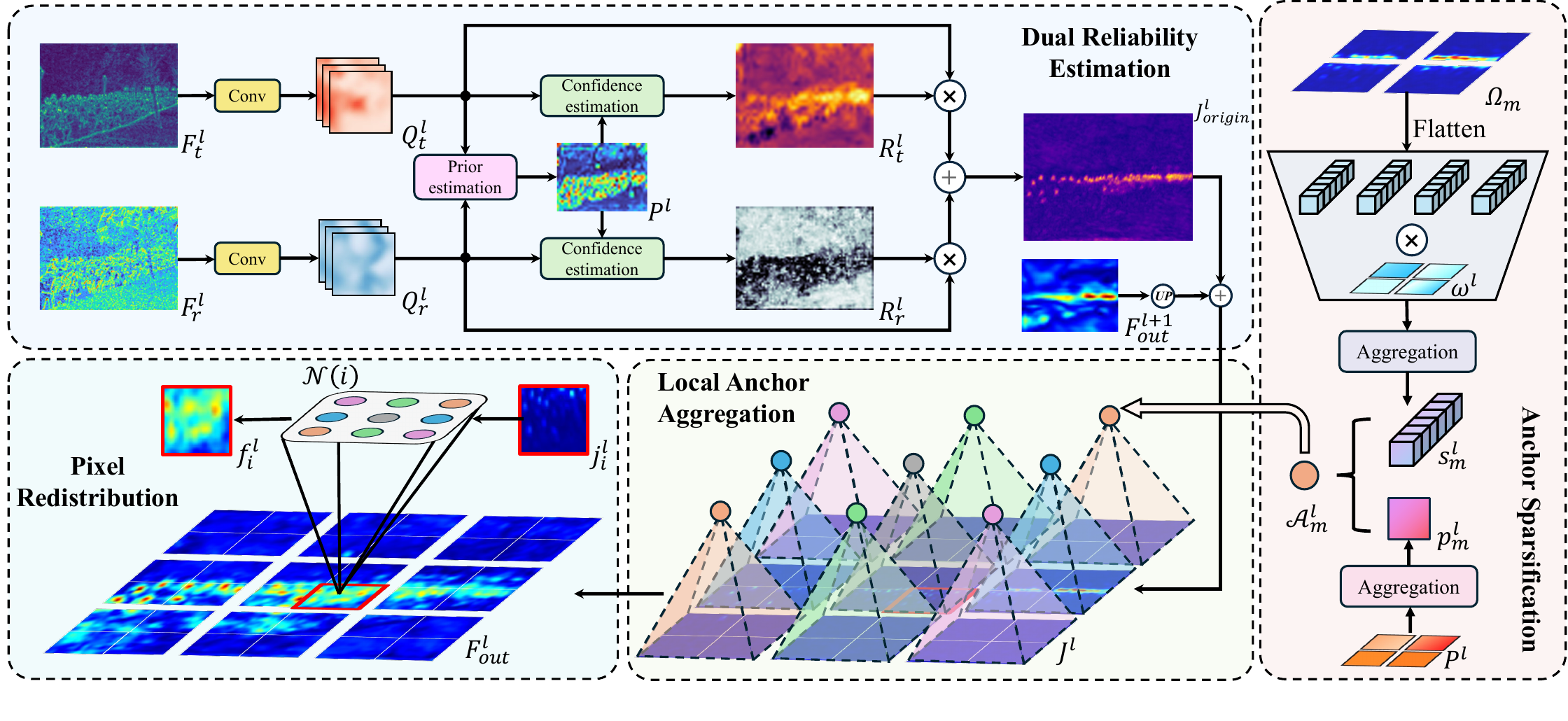}
    \vspace{-5mm}
   \captionsetup{font=footnotesize}
    \caption{Detailed architecture of the local anchor fusion module (LAFM). The module achieves feature fusion through three steps: (1) \textbf{Dual reliability estimation} (upper left): the joint crowd-aware prior $P^l$ is used to estimate modality-specific reliability maps ($R_r^l, R_t^l$), generating the weighted joint feature $J^l$; (2) \textbf{Anchor sparsification} (right): within the local window $\Omega_m$, features are aggregated into sparse anchors $\mathcal{A}_m^l$ to extract the crowd semantic prototype $s_m^l$ and the prior score $p_m^l$; (3) \textbf{Pixel redistribution} (lower left): each pixel $i$ interacts with its neighboring anchors $\mathcal{N}(i)$, remapping the aggregated global crowd semantics back to the pixel-level feature $f_i^l$.}
    \label{LAFM}
    \vspace{-2.6mm}
\end{figure*}

\section{Method}

\subsection{Overall Architecture}

To address the widely observed issues of local spatial misalignment and modality-quality imbalance in RGB-T crowd counting, we propose RACANet, a two-stage framework for RGB-T crowd counting. Given a pair of coarsely aligned visible-light and thermal infrared images, denoted by $I_r$ and $I_t$, respectively, RACANet employs a dual-branch PVTv2 backbone for feature extraction. The resulting multi-scale feature sets are formulated as $F_r = \{F_r^1, F_r^2, F_r^3, F_r^4\}, F_t = \{F_t^1, F_t^2, F_t^3, F_t^4\}$.

As illustrated in Fig.~\ref{fig:framework}, the training procedure of RACANet consists of two stages:

\textbf{(1) Cross-modal alignment pretraining:}
Before the formal counting training, RACANet learns cross-modal semantic correspondences through lightweight local soft matching and generates crowd-aware priors as spatial constraints for subsequent fusion.

\textbf{(2) Formal counting training:}
This stage inherits the pretrained backbone parameters and region priors. A local anchor fusion module (LAFM) guided by the crowd-aware prior is introduced at multiple feature scales to perform position-level modality reliability modeling. Finally, under a discrepancy-aware consistency constraint, the network yields multi-scale fused features $F_{\mathrm{out}} =
\{F_{\mathrm{out}}^1, F_{\mathrm{out}}^2, F_{\mathrm{out}}^3, F_{\mathrm{out}}^4\}$.

The fused features are then fed into the decoder to regress a highly robust crowd density map.

\subsection{Cross-modal Alignment Pretraining}

Due to sensor parallax and differences in imaging mechanisms, directly performing end-to-end fusion on locally misaligned bimodal features may easily introduce semantic confusion. Considering that cross-modal offsets in RGB-T scenes are typically local, we design a crowd-prior-guided local soft matching pretraining stage.

\paragraph{(1) Crowd-aware prior generation}

To reduce the computational complexity of subsequent processing and map features into a unified semantic space, we first apply low-dimensional projection to the features at the $l$-th stage:

\begin{equation}
Q_r^l = W_r^l(F_r^l), \qquad Q_t^l = W_t^l(F_t^l).
\end{equation}

Here, $Q_r^l, Q_t^l \in \mathbb{R}^{d^l \times H^l \times W^l}$ denote the low-dimensional projected features, and $W_r^l$ and $W_t^l$ are $1 \times 1$ projection convolutions. To avoid blindly aligning background regions lacking semantic information, we employ a lightweight convolutional branch $\phi_p^l$ to localize crowd-related regions that are valuable for alignment. Its input integrates the dual-modal features and their interaction differences:

\begin{equation}
P^l =
\sigma\!\left(
\phi_p^l \big(
[Q_r^l,\, Q_t^l,\, |Q_r^l - Q_t^l|,\, Q_r^l \odot Q_t^l]
\big)
\right),
\end{equation}

where $P^l \in \mathbb{R}^{1 \times H^l \times W^l}$ and $\sigma(\cdot)$ denotes the Sigmoid activation function. $P^l(i)$ represents the probability that spatial position $i$ contains crowd semantics, and we define it as the crowd-aware prior. We use the soft label $\tilde{P}^l$, obtained by applying Gaussian smoothing and downsampling to the ground-truth point annotations, as the supervision signal. The crowd-aware prior loss $\mathcal{L}_{cap}^l$ is computed by combining binary cross-entropy and Dice loss:

\begin{equation}
\mathcal{L}_{cap}^l
=
\mathrm{BCE}(P^l,\tilde{P}^l)
+
\bigl(1-\mathrm{Dice}(P^l,\tilde{P}^l)\bigr).
\end{equation}

\paragraph{(2) Prior-guided local soft matching}

After obtaining the crowd-aware prior, we perform bidirectional soft matching within a local window. Given the local offset set$ \Delta = \{(\Delta x,\Delta y)\mid \Delta x,\Delta y \in [-k_{\Delta},k_{\Delta}]\}$, where $k_{\Delta}$ denotes the radius of the local receptive field, for a position $i$ on the RGB feature map, the matching score between $i$ and the position $i+\delta$ in the local neighborhood of the thermal feature map is computed as:

\begin{equation}
e_{i,\delta}^{l,r\to t}
=
\frac{
Q_r^l(i)^{\top} Q_t^l(i+\delta)
}{\sqrt{d^l}}.
\end{equation}

The similarity distribution $\alpha_{i,\delta}^{l,r\to t}$ is then obtained by Softmax normalization within the local window:

\begin{equation}
\alpha_{i,\delta}^{l,r\to t}
=
\frac{
\exp\!\left(e_{i,\delta}^{l,r\to t}\right)
}{
\sum_{\delta' \in \Delta}
\exp\!\left(e_{i,\delta'}^{l,r\to t}\right)
}.
\end{equation}

Accordingly, the thermally soft-aligned feature at position $i$, denoted by $\tilde{Q}_t^l(i)$, is aggregated as:

\begin{equation}
\tilde{Q}_t^l(i)
=
\sum_{\delta \in \Delta}
\alpha_{i,\delta}^{l,r\to t}\, Q_t^l(i+\delta).
\end{equation}

Similarly, the reverse soft-aligned feature $\tilde{Q}_r^l(i)$ can be obtained. We then impose a weighted local consistency constraint using the crowd-aware prior $P^l$ to prevent the model from excessively focusing on background regions. The bidirectional alignment loss at the $l$-th stage is defined as:

\begin{equation}
\mathcal{L}_{align}^l
=
\frac{1}{\sum_i P_i^l + \varepsilon}
\sum_i P_i^l
\left(
\|Q_r^l(i)-\tilde{Q}_t^l(i)\|_1
+
\|Q_t^l(i)-\tilde{Q}_r^l(i)\|_1
\right).
\end{equation}

Here, $P_i^l$ serves as a positional weight, enabling the model to focus its alignment capability on crowd-related regions with higher semantic value. The total loss for the pretraining is defined as:

\begin{equation}
\mathcal{L}_{warm}
=
\sum_l
\left(
\mathcal{L}_{cap}^l + \mathcal{L}_{align}^l
\right).
\end{equation}

In this formulation, $\mathcal{L}_{cap}^l$ guides the network to learn the spatial distribution of crowd-related regions, while $\mathcal{L}_{align}^l$ is used to optimize local cross-modal semantic correspondences. After pretraining, the backbone parameters and the crowd-aware prior branch $\phi_p$ are transferred to the formal training stage.

\subsection{Local Anchor Fusion Module (LAFM)}

During the formal counting stage, we design a local anchor fusion module (LAFM) guided by the crowd-aware priors learned in the pretraining stage. As shown in Fig.~\ref{LAFM}, this module achieves low-complexity and interpretable cross-modal fusion through position-level modality reliability modeling, local anchor aggregation, and pixel-level semantic redistribution.

\subsubsection{Crowd-aware Prior and Reliability Estimation}

At stage $l$, we employ the prior branch $\phi_p^l$, obtained through pretraining, to generate the crowd-aware prior $P^l$. Since this prior is supervised by crowd point annotations, it is able to localize potential crowd-related regions and is therefore used as regional guidance for modality reliability modeling. Subsequently, $P^l$ is concatenated with the feature representations of the two modalities, respectively, and then fed into two independent lightweight convolutional branches, $\phi_r^l$ and $\phi_t^l$, to estimate the reliability of each modality at each spatial location:

\begin{equation}
R_r^l=\sigma\left(\phi_r^l([Q_r^l,P^l])\right),\qquad
R_t^l=\sigma\left(\phi_t^l([Q_t^l,P^l])\right),
\end{equation}

where $R_r^l, R_t^l \in \mathbb{R}^{1\times H^l\times W^l}$. The bimodal features are then fused through reliability weighting and combined with the output of the LAFM from the next stage to obtain the initial joint feature:

\begin{equation}
J^l=R_r^l\odot Q_r^l+R_t^l\odot Q_t^l+\operatorname{Upsample}(F_{out}^{l+1}).
\end{equation}
Here, the joint feature $J^l \in \mathbb{R}^{d^l \times H^l \times W^l}$ encodes the adaptively balanced contributions of different modalities at the current location, thereby providing a more stable intermediate representation for subsequent local prototype aggregation. Notably, for the final layer ($l = 4$), since there is no subsequent LAFM module, the joint feature is derived entirely from the initial fusion.

\subsubsection{Crowd-aware Local Anchor Sparsification}

To avoid the quadratic complexity of global attention mechanisms, we extract crowd prototypes based on local anchors. Specifically, we introduce an anchor coverage window of size $K_a\times K_a$ in the spatial domain, where $K_a=2k_{\Delta}+1$. The feature map of size $H^l\times W^l$ is divided into $M^l=\frac{H^l}{K_a}\times\frac{W^l}{K_a}$ non-overlapping local regions $\{\Omega_m\}_{m=1}^{M^l}$. When the boundary cannot be evenly divided by $K_a$, we apply zero padding to complete the boundary so as to ensure the integrity of the local window partitioning. For each region $\Omega_m$, we define a local anchor as $\mathcal{A}_m^l=\langle s_m^l,p_m^l\rangle$, where $s_m\in\mathbb{R}^{d^l}$ denotes the local crowd semantic prototype, and $p_m\in[0,1]$ is the regional crowd-prior score. For a pixel $i$, its weight with respect to the anchor is defined as:

\begin{equation}
w_i^l=P_i^l\cdot\frac{R_r^l(i)+R_t^l(i)}{2}.
\end{equation}

Subsequently, the two attributes of anchor $\mathcal{A}_m$ are computed through local aggregation as:

\begin{equation}
s_m^l=\frac{\sum_{i\in\Omega_m} w_i^l\, j_i^l}{\sum_{i\in\Omega_m} w_i^l+\varepsilon},
\qquad
p_m^l=\frac{1}{|\Omega_m|}\sum_{i\in\Omega_m} P_i^l.
\end{equation}

Here, $j_i^l$ denotes the vector representation of the joint feature $J^l$ at position $i$.
\begin{table*}[t]
\centering
\caption{Comparison of different methods on the RGBT-CC dataset.}
\vspace{-3mm}
\label{tab:rgbtcc_comparison}
\footnotesize
\setlength{\tabcolsep}{3.5mm}
\renewcommand{\arraystretch}{0.95}
\def\methodbox#1{\makebox[2.6cm][l]{#1}}
\resizebox{\textwidth}{!}{
\begin{tabular}{cccccccc}
\toprule
\methodbox{\textbf{Method}} & \textbf{Venue} & \textbf{Backbone} & \textbf{GAME0} & \textbf{GAME1} & \textbf{GAME2} & \textbf{GAME3} & \textbf{RMSE} \\ \midrule
\methodbox{MMCCN~\cite{mmccn}} & ACCV'20 & ResNet-50~\cite{ResNet} & 13.82 & 17.83 & 22.20 & 29.64 & 24.36 \\
\methodbox{BL+IADM~\cite{rgbtcc}} & CVPR'21 & VGG-19~\cite{vgg} & 15.61 & 19.95 & 24.69 & 32.89 & 28.18 \\
\methodbox{DEFNet~\cite{defnet}} & TITS'22 & VGG-16~\cite{vgg} & 11.90 & 16.08 & 20.19 & \underline{27.27} & 21.09 \\
\methodbox{CCANet~\cite{ccanet}} & TMM'23 & VGG-16~\cite{vgg} & 13.93 & 18.13 & 22.08 & 28.26 & 24.71 \\
\methodbox{CSA-Net~\cite{csanet}} & ESWA'23 & VGG-19~\cite{vgg} & 12.45 & 16.46 & 21.48 & 30.62 & 21.64 \\
\methodbox{CGINet~\cite{cginet}} & EAAI'23 & ConvNeXt~\cite{convnext} & 12.07 & 15.98 & 20.06 & 27.73 & 20.54 \\
\methodbox{MC3Net~\cite{mc3net}} & TITS'23 & ConvNeXt~\cite{convnext} & 11.47 & 15.06 & \underline{19.40} & 27.95 & 20.59 \\
\methodbox{MCN~\cite{mcn}} & ESWA'24 & PoolFormer~\cite{poolformer} & 11.56 & 15.92 & 20.16 & 28.06 & 19.02 \\
\methodbox{CFAF-Net~\cite{cfafnet}} & EAAI'24 & VGG-19~\cite{vgg} & 11.07 & 14.96 & 19.65 & 29.05 & \underline{18.83} \\
\methodbox{CrowdAlign~\cite{crowdalign}} & IVC'24 & VGG-19~\cite{vgg} & 11.07 & \underline{14.83} & 19.44 & 28.65 & 19.78 \\
\methodbox{GETANet~\cite{getanet}} & GRSL'24 & PVT~\cite{pvt} & 12.14 & 15.98 & \underline{19.40} & 28.61 & 22.17 \\
\methodbox{BGDFNet~\cite{bgdfnet}} & TIM'24 & VGG-16~\cite{vgg} & 11.00 & 15.04 & 19.86 & 29.72 & 19.05 \\
\methodbox{CSCA~\cite{csca}} & PR'25 & VGG-19~\cite{vgg} & 13.50 & 18.63 & 23.59 & 31.59 & 24.83 \\
\methodbox{MSPNet~\cite{mspnet}} & TCE'25 & IR-50~\cite{arcface} & 12.20 & 16.50 & 20.51 & 27.84 & 21.49 \\
\methodbox{MIANet~\cite{mianet}} & TITS'25 & VGG-19~\cite{vgg} & 11.97 & 15.65 & 19.93 & 27.54 & 22.17 \\
\methodbox{MISF-Net~\cite{misfnet}} & TMM'25 & VGG-16~\cite{vgg} & \underline{10.90} & 14.87 & 19.65 & 29.18 & 19.42 \\ \midrule
\methodbox{RACANet (Ours)} & - & PVTv2~\cite{pvtv2} & \textbf{10.18} & \textbf{14.19} & \textbf{18.12} & \textbf{25.33} & \textbf{18.13} \\ \bottomrule
\end{tabular}
}
\vspace{-2.1mm}
\end{table*}

\subsubsection{Anchor-guided Pixel Redistribution}

After obtaining the local anchors, LAFM redistributes the aggregated regional semantics back to pixel-level features. For a pixel $i$, it interacts only with the spatially neighboring candidate anchor set $\mathcal{N}(i)$. Specifically, $\mathcal{N}(i)$ is defined as a local anchor window of size $k_n\times k_n$, centered on the grid containing pixel $i$. The matching score between pixel $i$ and anchor $m$ is defined as:
\begin{equation}
\ell_{im}^l=
\frac{(j_i^l)^{T}s_m^l}{\sqrt{d^l}}
+\log(p_m^l+\varepsilon).
\end{equation}
Here, the first term measures the semantic similarity between the two, while the second term introduces a prior bias through the crowdness confidence $p_m^l$, thereby suppressing interference from regions with low crowd relevance. Subsequently, normalization is performed within the candidate set $\mathcal{N}(i)$ to obtain the adaptive attention weight $\alpha_{im}$:
\begin{equation}
\alpha_{im}^l=
\frac{\exp(\ell_{im}^l)}{\sum_{m'\in\mathcal{N}(i)}\exp(\ell_{im'}^l)},
\qquad
m\in\mathcal{N}(i).
\end{equation}
Finally, pixel $i$ re-aggregates the semantic prototypes from its local neighborhood, thereby completing the cross-modal semantic redistribution of features:
\begin{equation}
f_i^l=\sum_{m\in\mathcal{N}(i)}\alpha_{im}^ls_m^l.
\end{equation}
The redistributed features $\{f_i^l\}_{i=1}^{H^lW^l}$ are then rearranged into $F_{\mathrm{out}}^l$, which is passed as the fused output of the current stage to the density regression decoder.
\begin{table*}[t]
\centering
\caption{Comparison of different methods on the Drone-RGBT dataset.}
\vspace{-3mm}
\label{tab:dronergbt_comparison}
\footnotesize
\setlength{\tabcolsep}{3.5mm}
\renewcommand{\arraystretch}{0.95}
\def\methodbox#1{\makebox[2.6cm][l]{#1}}
\resizebox{\textwidth}{!}{
\begin{tabular}{cccccccc}
\toprule
\methodbox{\textbf{Method}} & \textbf{Venue} & \textbf{Backbone} & \textbf{GAME0} & \textbf{GAME1} & \textbf{GAME2} & \textbf{GAME3} & \textbf{RMSE} \\ \midrule
\methodbox{MMCCN~\cite{mmccn}} & ACCV'20 & ResNet-50~\cite{ResNet} & 7.27 & - & - & - & 11.45 \\
\methodbox{BL+IADM~\cite{rgbtcc}} & CVPR'21 & VGG-19~\cite{vgg} & 9.70 & 12.04 & 15.31 & 20.31 & 15.01 \\
\methodbox{DEFNet~\cite{defnet}} & TITS'22 & VGG-16~\cite{vgg} & 7.89 & 9.60 & 11.96 & 15.34 & 12.88 \\
\methodbox{CGINet~\cite{cginet}} & EAAI'23 & ConvNeXt~\cite{convnext} & 8.37 & 9.97 & 12.34 & 15.51 & 13.45 \\
\methodbox{MC3Net~\cite{mc3net}} & TITS'23 & ConvNeXt~\cite{convnext} & 7.33 & - & - & - & 11.17 \\
\methodbox{CrowdAlign~\cite{crowdalign}} & IVC'24 & VGG-19~\cite{vgg} & 7.03 & - & - & - & 10.96 \\
\methodbox{GETANet~\cite{getanet}} & GRSL'24 & PVT~\cite{pvt} & 8.44 & 10.01 & 12.75 & 15.83 & 13.99 \\

\methodbox{C4-MIM~\cite{c4mim}} & CAIS'24 & VGG-19~\cite{vgg} & 6.72 & - & - & - & 10.58 \\
\methodbox{BMCC~\cite{bmcc}} & ECCV'24 & VGG-19~\cite{vgg} \& ViT~\cite{vit} & \underline{6.20} & - & - & - & \underline{10.40} \\
\methodbox{CSCA~\cite{csca}} & PR'25 & VGG-19~\cite{vgg} & 9.51 & 12.12 & 15.84 & 21.57 & 15.19 \\
\methodbox{MIANet~\cite{mianet}} & TITS'25 & VGG-19~\cite{vgg} & 6.74 & \underline{8.64} & \underline{11.49} & 16.31 & 10.58 \\ 
\methodbox{CMFX~\cite{cmfx}} & NN'25 & VGG-19~\cite{vgg} & 6.75 & 8.88 & 11.87 & \underline{14.69} & 11.05 \\ \midrule
\methodbox{RACANet (Ours)} & - & PVTv2~\cite{pvtv2} & \textbf{5.23} & \textbf{6.69} & \textbf{8.78} & \textbf{12.08} & \textbf{8.18} \\ \bottomrule
\end{tabular}
}
\vspace{-3.5mm}
\end{table*}
\subsection{Discrepancy-aware Reliability Consistency Constraint}

Although independent reliability modeling is more consistent with real-world scenarios, if the reliability maps $R_r^l$ and $R_t^l$ are left completely unconstrained, the network may learn unreasonable modality biases in semantically consistent regions. To address this issue, we propose a discrepancy-aware consistency constraint. Specifically, a cross-modal discrepancy map $D^l$ is first constructed using the absolute difference between modal features:

\begin{equation}
D^l=\sigma\!\left(\phi_d^l\!\left(\lvert Q_r^l-Q_t^l\rvert\right)\right),\qquad
D^l\in\mathbb{R}^{1\times H^l\times W^l}.
\end{equation}

Here, $D_i^l\in[0,1]$ denotes the degree of cross-modal discrepancy at position $i$. Based on this, the consistency loss is defined as:

\begin{equation}
\mathcal{L}_{cons}^l
=
\frac{1}{H^lW^l}\sum_{i=1}^{H^lW^l}(1-D_i^l)\,\lvert R_r^l(i)-R_t^l(i)\rvert.
\end{equation}

This loss imposes a strong consistency constraint when $D_i^l$ is small, namely in regions where the two modalities are semantically consistent, thereby preventing the introduction of unnecessary noise. In contrast, the constraint is relaxed in highly discrepant regions, thus preserving modality-specific characteristics and complementary advantages.

\subsection{Density Map Regression}

After obtaining the fused features from the four stages, $\{F_{\mathrm{out}}^l\}_{l=1}^{4}$, the network generates the predicted density map through a progressive multi-scale decoder. Specifically, low-resolution features are upsampled, then channel-wise concatenated with high-resolution features, followed by convolutional fusion. Finally, a regression head composed of batch normalization and ReLU activation outputs a single-channel crowd density map $\hat{Y}\in\mathbb{R}^{H\times W}$.

To supervise the density regression process, we adopt the Bayesian loss as the main counting loss $\mathcal{L}_{cnt}$. Let $\mathcal{P}=\{p_1,p_2,\ldots,p_N\}$ denote the set of ground-truth point distributions, where $p_i\in\mathbb{R}^{H\times W}$ represents the spatial probability distribution of the $i$-th target point. The corresponding predicted count is defined as:

\begin{equation}
\hat{c}_i=\sum_{j=1}^{HW}p_i[j]\cdot \hat{Y}[j].
\end{equation}

For each ground-truth target point, the expected true count is set to $t_i=1$. The main counting loss is then defined as:

\begin{equation}
\mathcal{L}_{cnt}
=
\frac{1}{|\mathcal{P}|}\sum_{i=1}^{|\mathcal{P}|}\lvert t_i-\hat{c}_i\rvert.
\end{equation}

Compared with the conventional pixel-wise mean squared error, the Bayesian loss  integrates the probabilistic prior of target points with pixel-level density map supervision, thereby providing a more robust optimization objective for the density regression task. Finally, the overall optimization objective $\mathcal{L}_{total}$ is defined as:

\begin{equation}
\mathcal{L}_{total}
=
\mathcal{L}_{cnt}
+
\lambda_{cons}\sum_{l=1}^{4}\mathcal{L}_{cons}^{\,l},
\end{equation}

where $\lambda_{cons}$ is a balancing coefficient that controls the strength of the consistency constraint.

\section{Experiment}

\subsection{Datasets}

We evaluate our method on two mainstream RGB-T crowd counting benchmarks:

\textbf{RGBT-CC}: This dataset contains 2,030 image pairs with 138,389 annotated instances. It covers a variety of crowded scenes, such as streets and shopping malls, under diverse illumination conditions, and exhibits evident cross-modal appearance discrepancies and substantial scene complexity.

\textbf{Drone-RGBT}: This dataset is designed for unmanned aerial vehicle (UAV) viewpoints and contains 3,607 image pairs with 175,698 annotated instances. It includes multiple open scenes such as campuses and parking lots, and is characterized by large scale variations, strong perspective distortion, and complex local occlusions.

\subsection{Implementation Details}

All experiments are conducted on a single NVIDIA RTX 4090 GPU. The backbone network adopts PVTv2-B3 pretrained on ImageNet-1k. The learning rates for the pretraining stage and the main training stage are set to $1\times 10^{-5}$ and $1\times 10^{-4}$, respectively. The batch size is set to 16 for both stages, and the numbers of training epochs are 30 and 300, respectively. For the local modeling parameters, the anchor aggregation scale $K_a$ is uniformly set to 3, while the pixel redistribution neighborhood size $k_n$ is set to 5 and 3 for the RGBT-CC and Drone-RGBT datasets, respectively. The consistency loss coefficient $\lambda_{cons}$ is set to 0.1. During training, the input images are randomly cropped to $256\times256$ and augmented with random horizontal flipping. In addition, geometry-adaptive Gaussian kernels are used to generate soft labels and ground-truth density maps.

\subsection{Evaluation Metrics}

We adopt the standard crowd counting metrics, namely Grid Average Mean Absolute Error (GAME) and Root Mean Square Error (RMSE), for evaluation. GAME$(L)$ partitions an image into $4^L$ non-overlapping grids and computes the sum of absolute errors over all regions:

\begin{equation}
GAME(L)=\frac{1}{N}\sum_{n=1}^{N}\sum_{i=1}^{4^L}\left|P_n^i-G_n^i\right|.
\end{equation}

Here, $N$ denotes the total number of test images, and $P_n^i$ and $G_n^i$ denote the predicted and ground-truth crowd counts in the $i$-th grid of the $n$-th image, respectively. When $L=0$, GAME$(0)$ reduces to the global Mean Absolute Error (MAE). RMSE is used to evaluate global performance and is defined as:

\begin{equation}
RMSE=\sqrt{\frac{1}{N}\sum_{n=1}^{N}(P_n-G_n)^2}.
\end{equation}

Here, $P_n$ and $G_n$ denote the predicted and ground-truth crowd counts of the $n$-th image, respectively.
\begin{table*}[t]
\centering
\caption{Ablation study of different components on RGBT-CC and Drone-RGBT datasets. The checkmark indicates the module or loss is enabled.}
\vspace{-3mm}
\label{tab:ablation}
\footnotesize
\setlength{\tabcolsep}{3.5mm}
\renewcommand{\arraystretch}{0.95}
\resizebox{\textwidth}{!}{
\begin{tabular}{ccccccccccc}
\toprule
\multirow{2}{*}{\textbf{Baseline}} & \multirow{2}{*}{\textbf{Pretraining}} & \multirow{2}{*}{\textbf{LAFM}} & \multirow{2}{*}{\textbf{$\mathcal{L}_{\mathrm{cons}}$}} & \multicolumn{2}{c}{\textbf{RGBT-CC}} & \multicolumn{2}{c}{\textbf{Drone-RGBT}} & \multirow{2}{*}{\textbf{Params(M)}} & \multirow{2}{*}{\textbf{GFLOPs}} & \multirow{2}{*}{\textbf{FPS}} \\
\cmidrule(lr){5-6} \cmidrule(lr){7-8}
& & & & \textbf{GAME0} & \textbf{RMSE} & \textbf{GAME0} & \textbf{RMSE} & & & \\
\midrule
\checkmark & & & & 11.85 & 20.84 & 7.06 & 12.43 & \multirow{2}{*}{94.45} & \multirow{2}{*}{51.60} & \multirow{2}{*}{29.49} \\
\checkmark & \checkmark & & & 11.54 & 19.62 & 6.39 & 10.55 & & & \\
\cmidrule(lr){9-11}
\checkmark & & \checkmark & & 11.36 & 19.78 & 6.03 & 9.20 & \multirow{4}{*}{90.91} & \multirow{4}{*}{43.09} & \multirow{4}{*}{37.25} \\
\checkmark & & \checkmark & \checkmark & 10.89 & 19.01 & 5.91 & 8.95 & & & \\
\checkmark & \checkmark & \checkmark & & 10.55 & 18.71 & 5.51 & 8.53 & & & \\
\checkmark & \checkmark & \checkmark & \checkmark & \textbf{10.18} & \textbf{18.13} & \textbf{5.23} & \textbf{8.18} & & & \\
\bottomrule
\end{tabular}
}
\vspace{-3mm}
\end{table*}
\subsection{Comparison with Other Methods}

As shown in Tables~\ref{tab:rgbtcc_comparison} and ~\ref{tab:dronergbt_comparison}, compared with various state-of-the-art RGB-T crowd counting methods, RACANet achieves the best performance across all evaluation metrics.

Specifically, on the RGBT-CC dataset, RACANet outperforms the strongest competing method for each metric, reducing GAME$(0)$ by 0.72 (6.61\%) over MISF-Net and RMSE by 0.70 (3.72\%) over CFAF-Net. Furthermore, it also demonstrates clear advantages on GAME$(1)$ to GAME$(3)$, which reflect local spatial localization accuracy. Compared with DEFNet, the GAME$(3)$ score of RACANet decreases from 27.27 to 25.33. These results indicate that the proposed model not only improves global counting accuracy but also recovers the spatial distribution structure more precisely.

On the Drone-RGBT dataset, under the UAV viewpoint with severe scale variation, RACANet achieves GAME$(0)$ and RMSE scores of 5.23 and 8.18, respectively, surpassing the best competing method BMCC by 0.97 (15.65\%) and 2.22 (21.35\%). Notably, compared with BMCC, which adopts a two-stage VGG-ViT hybrid architecture, RACANet achieves superior performance with a relatively lightweight PVTv2 backbone, demonstrating the effectiveness of the pretraining stage and the LAFM module.

Overall, RACANet effectively overcomes local cross-modal misalignment and exploits complementary information from the two modalities while maintaining a simple and efficient architecture, thereby achieving state-of-the-art accuracy.

\subsection{Ablation Studies}

\subsubsection{Ablation on the Proposed Components}

To validate the effectiveness of the key components of RACANet, we conducted ablation studies, and the results are reported in Table~\ref{tab:ablation}. To ensure comparability, the baseline model adopts a dual-branch PVTv2 as the feature extraction backbone, employs conventional cross-attention with convolutional layers for cross-modal fusion, and removes the pretraining stage. After introducing the cross-modal alignment pretraining on top of the baseline, the RMSE on RGBT-CC and Drone-RGBT improves by 1.22 (5.9\%) and 1.88 (15\%), respectively, demonstrating that the local cross-modal alignment prior can provide a more reliable initialization for subsequent counting learning. As modules such as the LAFM are progressively introduced, the model performance continues to improve. Meanwhile, compared with conventional cross-attention, the proposed LAFM reduces the parameter count from 94.45M to 90.91M and the GFLOPs from 51.60 to 43.09, demonstrating the lightweight and efficient design of the proposed method.

\subsubsection{Ablation Study on Pretraining Losses}

We further conducted an ablation analysis of the crowd-aware prior supervision loss $\mathcal{L}_{cap}$ and the bidirectional feature alignment loss $\mathcal{L}_{align}$ in the pretraining stage on the RGBT-CC and Drone-RGBT datasets. The results are reported in Table~\ref{tab:pretrain_ablation}. Here, “RACANet w/o pretraining” denotes the full model trained without the pretraining stage. It can be observed that the model achieves the best performance on both datasets when $\mathcal{L}_{cap}$ and $\mathcal{L}_{align}$ are introduced simultaneously. This indicates that the former enhances the model's ability to perceive crowd-related regions, while the latter helps improve the consistency and robustness of cross-modal feature representations.
\begin{table}[t]
\centering
\caption{Ablation study of the pretraining stage on RGBT-CC and Drone-RGBT datasets.}
\vspace{-3mm}
\label{tab:pretrain_ablation}
\footnotesize
\renewcommand{\arraystretch}{0.95}
\begin{tabular*}{\linewidth}{@{\hspace{4pt}\extracolsep{\fill}}ccccc@{\hspace{4pt}}}
\toprule
\multirow{2}{*}{\textbf{Method}} & \multicolumn{2}{c}{\textbf{RGBT-CC}} & \multicolumn{2}{c}{\textbf{Drone-RGBT}} \\ \cmidrule(lr){2-3} \cmidrule(lr){4-5}
 & \textbf{GAME0} & \textbf{RMSE} & \textbf{GAME0} & \textbf{RMSE} \\ \midrule
RACANet w/o pretraining & 10.89 & 19.01 & 5.91 & 8.95 \\
RACANet w/o $\mathcal{L}_{\mathrm{align}}$ & 10.84 & 18.90 & 5.74 & 8.82 \\
RACANet w/o $\mathcal{L}_{\mathrm{cap}}$ & \underline{10.57} & \underline{18.30} & \underline{5.39} & \underline{8.34} \\
RACANet & \textbf{10.18} & \textbf{18.13} & \textbf{5.23} & \textbf{8.18} \\ \bottomrule
\end{tabular*}
\vspace{-3mm}
\end{table}
\begin{table}[t]
\centering
\caption{Ablation study of the anchor window size $K_a$ on RGBT-CC and Drone-RGBT datasets.}
\vspace{-3mm}
\label{tab:anchor_k_ablation}
\footnotesize
\renewcommand{\arraystretch}{0.95}
\begin{tabularx}{\linewidth}{@{} *{5}{>{\centering\arraybackslash}X} @{}}
\toprule
\multirow{2}{*}{\textbf{$K_a$}} & \multicolumn{2}{c}{\textbf{RGBT-CC}} & \multicolumn{2}{c}{\textbf{Drone-RGBT}} \\ \cmidrule(lr){2-3} \cmidrule(lr){4-5}
 & \textbf{GAME0} & \textbf{RMSE} & \textbf{GAME0} & \textbf{RMSE} \\ \midrule
1 & 10.82 & 18.99 & 5.89 & 8.75 \\
3 & \textbf{10.18} & \textbf{18.13} & \textbf{5.23} & \textbf{8.18} \\
5 & 10.56 & \underline{18.31} & \underline{5.36} & \underline{8.21} \\
7 & \underline{10.54} & 18.45 & 5.53 & 8.49 \\
9 & 10.86 & 18.84 & 6.04 & 8.87 \\ \bottomrule
\end{tabularx}
\vspace{-4mm}
\end{table}
\begin{figure*}[t]
    \centering
    \includegraphics[width=0.95\textwidth]{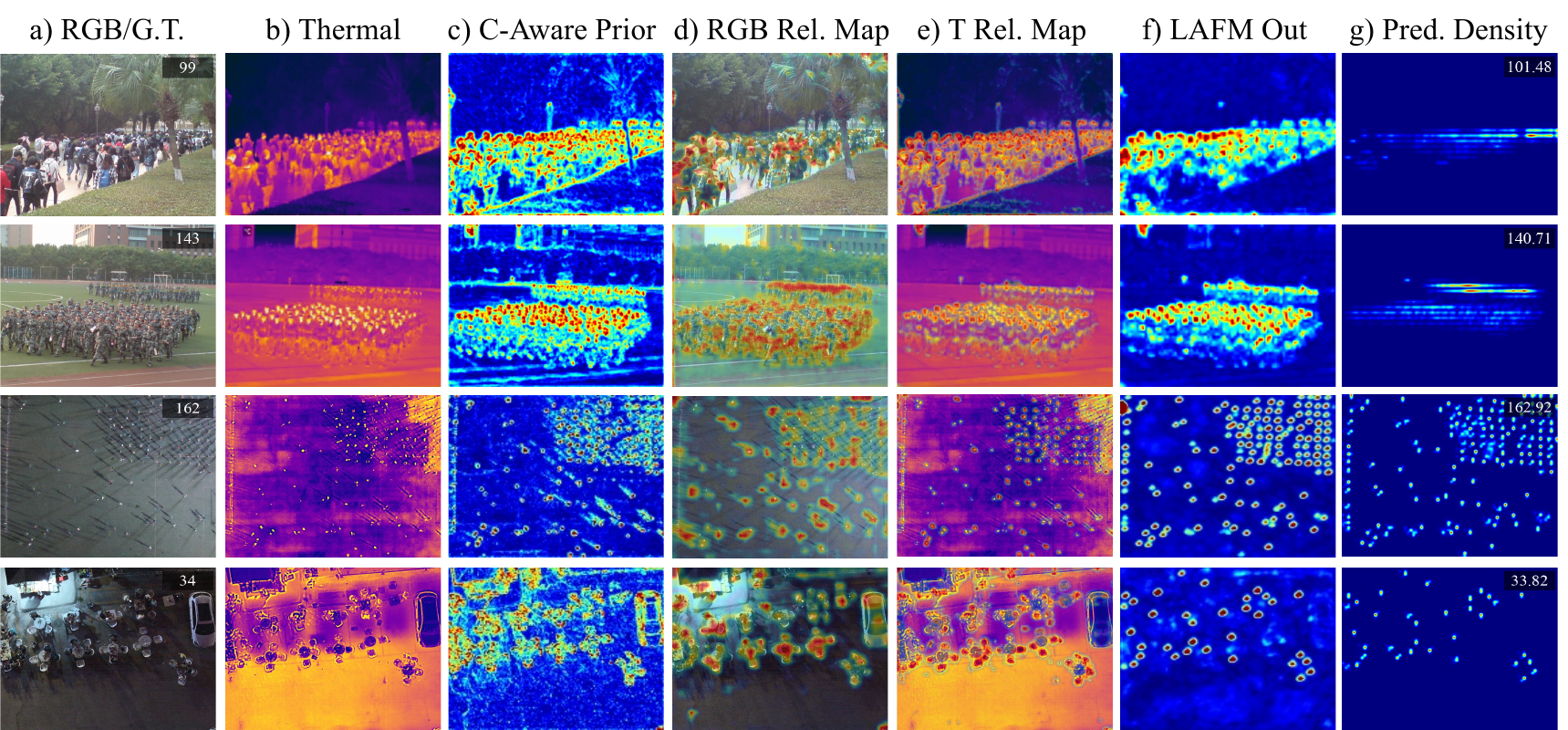}
    \vspace{-2.5mm}
    \captionsetup{font=footnotesize}
    \caption{Visualization results of the proposed RACANet under various complex scenarios. From left to right: (a) RGB image and the ground-truth count; (b) thermal infrared image; (c) crowd-aware prior $P^l$; (d) reliability map of the RGB modality, $R_r$, where highlighted regions indicate higher confidence of this modality at the corresponding local positions; (e) reliability map of the thermal modality, $R_t$; (f) output features of the LAFM; (g) predicted density map and the estimated crowd count. The visualization results show that RACANet can adaptively assess modality reliability and accurately estimate crowd density under conditions of insufficient illumination or thermal noise interference.}
    \label{fig:visual}
    \vspace{-2.1mm}
\end{figure*}
\subsubsection{Ablation Study on Hyperparameters}

The anchor aggregation scale $K_a$ and the pixel redistribution neighborhood size $k_n$ in the local anchor fusion module (LAFM) determine the receptive-field range of local cross-modal interactions, while the balancing coefficient $\lambda_{cons}$ controls the strength of the discrepancy-aware consistency constraint. With all other settings kept unchanged, we conducted an ablation analysis of these three key hyperparameters.

\paragraph{Local Anchor Aggregation Scale ($K_a$)}
As shown in Table~\ref{tab:anchor_k_ablation}, when $K_a$ is small, the local window covers only a limited range, making it difficult for the generated anchor prototypes to capture sufficient semantic information. In contrast, when $K_a$ is too large, the aggregation range exceeds the reasonable assumption of local offsets, leading to performance degradation. Experimental results show that the model achieves the best aggregation capability on both datasets when $K_a=3$.
\begin{table}[t]
\centering
\caption{Ablation study of the neighborhood window size $k_n$ on
RGBT-CC and Drone-RGBT datasets.}
\vspace{-3mm}
\label{tab:neighbor_k_ablation}
\footnotesize
\renewcommand{\arraystretch}{0.95}
\begin{tabularx}{\linewidth}{@{} *{5}{>{\centering\arraybackslash}X} @{}}
\toprule
\multirow{2}{*}{\textbf{$k_n$}} & \multicolumn{2}{c}{\textbf{RGBT-CC}} & \multicolumn{2}{c}{\textbf{Drone-RGBT}} \\ \cmidrule(lr){2-3} \cmidrule(lr){4-5}
 & \textbf{GAME0} & \textbf{RMSE} & \textbf{GAME0} & \textbf{RMSE} \\ \midrule
1 & 10.50 & 18.57 & 5.40 & 8.53 \\
3 & 10.33 & 18.54 & \textbf{5.23} &\textbf{8.18}  \\
5 & \textbf{10.18} & \textbf{18.13} & \underline{5.38} & 8.44 \\
7 & \underline{10.27} & \underline{18.33} & 5.52 & \underline{8.26} \\ \bottomrule
\end{tabularx}
\vspace{-2mm}
\end{table}

\paragraph{Pixel Redistribution Neighborhood Size ($k_n$)}
As shown in Table~\ref{tab:neighbor_k_ablation}, when $k_n$ is set too small, each pixel can only receive limited local anchor information, which restricts the utilization of cross-modal complementary features. In contrast, when $k_n$ is too large, excessive surrounding smoothing information is introduced, thereby weakening the discriminability of local features. In RGBT-CC, where crowd distributions in street-view and indoor scenes are denser and more continuous, a larger reassignment neighborhood ($k_n=5$) can provide pixels with richer crowd-context semantics. By contrast, in the Drone-RGBT dataset, crowd targets under aerial drone views are usually extremely small and exhibit drastic scale variations; thus, a smaller neighborhood ($k_n=3$) can effectively prevent the discriminative features of small-scale targets from being overly smoothed by the surrounding background.
\begin{table}[t]
\centering
\caption{Ablation study of the consistency loss weight $\lambda_{\mathrm{cons}}$ on the RGBT-CC dataset.}
\vspace{-3mm}
\label{tab:lambda_cons_ablation}
\footnotesize
\renewcommand{\arraystretch}{0.95}
\begin{tabular*}{\linewidth}{@{\hspace{6pt}\extracolsep{\fill}}cccccc@{\hspace{6pt}}}
\toprule
\textbf{$\lambda_{\mathrm{cons}}$} & \textbf{GAME0} & \textbf{GAME1} & \textbf{GAME2} & \textbf{GAME3} & \textbf{RMSE} \\ \midrule
0.001 & 10.52 & 14.76 & \underline{18.83} & 26.77 & 18.54 \\
0.01  & 10.43 & 15.01 & 18.94 & 26.39 & \underline{18.43} \\
0.1   & \textbf{10.18} & \textbf{14.19} & \textbf{18.12} & \textbf{25.33} & \textbf{18.13} \\
1     & \underline{10.26} & \underline{14.72} & 19.11 & \underline{26.02} & 18.59 \\
10    & 10.93 & 15.49 & 19.23 & 27.64 & 19.02 \\ \bottomrule
\end{tabular*}
\vspace{-4mm}
\end{table}

\paragraph{Consistency Constraint Coefficient ($\lambda_{cons}$)}
As shown in Table~\ref{tab:lambda_cons_ablation}, when $\lambda_{cons}$ is too small, the proposed constraint is insufficient to effectively guide the two modalities to learn stable reliability distributions in semantically consistent regions. In contrast, when $\lambda_{cons}$ is set too large, overly strong restrictions are imposed on regions with genuine modality differences, thereby weakening modality complementarity. The experimental results indicate that the model achieves the best performance when $\lambda_{cons} = 0.1$.

\begin{table}[t]
\centering
\caption{Comparison of different backbones on the Drone-RGBT dataset.}
\vspace{-3mm}
\label{tab:backbone_comparison}
\resizebox{\linewidth}{!}{
\begin{tabular}{lccccc}
\toprule
\textbf{Backbone} & \textbf{GAME0} & \textbf{RMSE} & \textbf{Params(M)} & \textbf{GFLOPs} & \textbf{FPS} \\ \midrule
VGG-19~\cite{vgg} & 6.07 & 10.14 & \textbf{41.55} & 108.73 & \textbf{56.87} \\
ConvNeXt-T~\cite{convnext} & \underline{5.89} & \underline{9.50} & \underline{57.15} & \underline{43.91} & \underline{56.81} \\
Swin-Transformer-S~\cite{swintransformer} & 6.09 & 10.27 & 99.18 & 63.95 & 30.17 \\
SegFormer-MiT-B3~\cite{segformer} & 5.91 & 9.64 & 91.06 & 45.14 & 30.84 \\
PVTv2-B3~\cite{pvtv2} & \textbf{5.23} & \textbf{8.18} & 90.91 & \textbf{43.09} & 37.25 \\ \bottomrule
\end{tabular}
}
\vspace{-0.45cm}
\end{table}
\subsubsection{Comparison with Different Backbones}

To verify the rationality of the selected backbone, under the unified setting of using cross-modal alignment pretraining, we compare PVTv2-B3 with several mainstream visual backbones, including the CNN-based architectures VGG-19 and ConvNeXt-T, as well as the Transformer-based architectures Swin-Transformer-S and SegFormer-MiT-B3. Table~\ref{tab:backbone_comparison} reports the results in terms of the number of parameters, computational complexity, inference speed, and counting accuracy. The experimental results show that PVTv2-B3 achieves a better balance between performance and computational cost.

\subsection{Qualitative Analysis}

Figure~\ref{fig:visual} presents qualitative visualization results of RACANet under different scenarios. First, the crowd-aware prior map in (c) intuitively demonstrates its ability to accurately capture crowd regions. Second, the reliability maps within the LAFM module in (d) and (e), together with the output feature in (f), illustrate the process of adaptive feature fusion. In the thermal interference scenario (Row 2), the RGB reliability map dominates due to its clear visual boundaries. In contrast, in the nighttime scenes (Rows 3 and 4), the thermal reliability map is significantly superior to the RGB one in dark regions, reflecting the network's adaptive exploitation of modality complementarity. Compared with the crowd-aware prior, the output feature of LAFM is more refined. In both dense scenes (Rows 1 and 2) and sparse scenes (Rows 3 and 4), this feature effectively suppresses background interference from trees, vehicles, and other irrelevant objects, indicating that the LAFM module helps extract crowd-related representations. Finally, the predicted density maps in (g) remain highly consistent with the ground truth in (a) in terms of both spatial distribution and numerical estimation, validating the superiority of RACANet in addressing local spatial misalignment and imbalanced modality quality.

\vspace{2pt}
\section{Conclusion}
In this paper, we propose RACANet, a two-stage framework for RGB-T crowd counting, to address local cross-modal misalignment and imbalanced modality reliability. By introducing a lightweight cross-modal alignment pretraining stage, the network explicitly learns local spatial correspondences, thereby providing stable crowd-aware priors for subsequent feature fusion. On this basis, we further design a local anchor fusion module (LAFM), which combines position-level reliability estimation with a semantic redistribution mechanism to achieve efficient and interpretable cross-modal feature integration. In addition, we introduce a discrepancy-aware reliability consistency constraint to further balance modality complementarity and consistency. Experimental results on the RGBT-CC and Drone-RGBT datasets demonstrate that RACANet outperforms existing state-of-the-art methods in both counting accuracy and robustness, validating its application potential in complex scenarios.

\clearpage


\bibliographystyle{ACM-Reference-Format}
\bibliography{sample-base}

\appendix

\end{document}